\documentclass[conference]{IEEEtran}

\usepackage{hyperref}
\usepackage{graphicx}
\usepackage{multirow}
\usepackage{amsmath,amsthm,amsfonts}
\usepackage{amssymb}
\usepackage{bbm}

\theoremstyle{definition}
\newtheorem{definition}{Definition}

\usepackage{xcolor}

\begin{document}



%\twocolumn[

% \aistatstitle{Closed-Loop View of the Regulation of AI: \\ Equal Impact across Repeated Interactions}

% \aistatsauthor{ Quan Zhou \And Ramen Ghosh \And  Robert Shorten \And Jakub Marecek}

% \aistatsaddress{ Imperial College London \And  University College Dublin \And Czech Technical University in Prague } ]

\title{Closed-Loop View of the Regulation of AI: \\ Equal Impact across Repeated Interactions}

% \author{
% \IEEEauthorblockN{Quan Zhou}
% \IEEEauthorblockA{Imperial College London}
% \and
% \IEEEauthorblockN{Ramen Ghosh}
% \IEEEauthorblockA{\\
% Atlantic Technological University}

% \and
% \IEEEauthorblockN{Robert Shorten}
% \IEEEauthorblockA{Imperial College London}
% \and
% \IEEEauthorblockN{Jakub Mare\v{c}ek}
% \IEEEauthorblockA{Czech Technical University in Prague}
% }

\author{
\IEEEauthorblockN{Quan Zhou\IEEEauthorrefmark{1},
Ramen Ghosh\IEEEauthorrefmark{2},
Robert Shorten\IEEEauthorrefmark{1}, and
Jakub Mare\v{c}ek\IEEEauthorrefmark{3}}
\IEEEauthorblockA{\IEEEauthorrefmark{1}
Imperial College London}
\IEEEauthorblockA{\IEEEauthorrefmark{2}
Atlantic Technological University}
\IEEEauthorblockA{\IEEEauthorrefmark{3}Czech Technical University in Prague}
}

\maketitle

\begin{abstract}
There has been much recent interest in the regulation of AI. We argue for a view based on civil-rights legislation, built on the notions of equal treatment and equal impact. In a closed-loop view of the AI system and its users, the equal treatment concerns one pass through the loop. Equal impact, in our view, concerns the long-run average behaviour across repeated interactions. In order to establish the existence of the average and its properties, one needs to study the ergodic properties of the closed-loop and, in particular, its unique stationary measure.   
\end{abstract}

\section{ Introduction }

There has been considerable interest in the regulation of artificial intelligence (AI), recently. 
It is increasingly recognised that so-called high-risk applications of AI, such as in human resources, retail banking, or within public schools, be it admissions or assessment, cannot be served by black-box AI systems with no human control \cite{bringas2022fairness}, predominantly due to concerns for protected human rights. 
A great many reports and research have revealed the danger of AI systems violating fairness in predicting which areas need patrolling \cite{courtland2018bias}, criminal-risk assessment \cite{angwin2016machine}, discriminatory behavior in advertising and recruiting algorithms for people with disabilities  \cite{nugent2021recruitment,guo2020toward}, search engine reinforcing racism \cite{noble2018algorithms}; and the threat of breaching privacy \cite{nguyen2021security,sun2020machine}.
To cope with the challenges of AI, leading technology companies have issued AI principles of their own and developed software tools geared towards fairness and explainbility of AI, such as AIF360 \cite{aif360-oct-2018} of IBM, %, AIX360 \cite{aix360-sept-2019}
SHAP \cite{SHAP} of Microsoft. %FairLearn \cite{bird2020fairlearn}, 
%and Facets \cite{facets}, What-If tool \cite{What-If} of Google.
In a broader context, it is not clear \cite{dobbe2021hard}, however, how to phrase even the desiderata for the regulation of AI. 

Here, we suggest that the desiderata could be the same as in the Civil Rights Act of 1964 and much of the subsequent civil-right legislation world-wide: equal treatment and equal impact.
At the same time, we point out that these desiderata could be in conflict \cite{binns2020apparent,zhao2019inherent}. 
The Ricci v. DeStefano, 557 U.S. 557 (2009) labour law case has demonstrated the practical differences between them, where the city of New Haven has declined to promote city firefighters based on the same test, which, shows a disproportionate pass rate for a certain race, as to the fear of valiating Title VII of the Civil Right Act of 1964 \cite{mcginley2011ricci}. The use of the same test conducts the principle of equal treatment, while the disparate pass rates and possibly contrasting promotion results do not comply with equal impact.

Let us illustrate the conflict with another example of a system that performs credit-risk estimation in a consumer-credit company.
%\footnote{See also Consumer Financial Protection Circular 2022-03: \url{https://www.consumerfinance.gov/compliance/circulars/circular-2022-03-adverse-action-notification-requirements-in-connection-with-credit-decisions-based-on-complex-algorithms/} for a detailed discussion of its meaning for AI systems.}
In the US, this is regulated by the Equal Credit Opportunity Act of 1974, but the example applies equally well to other countries. 
Imagine a situation where the the credit decision is uniform: everyone who has not defaulted on any loan is approved a credit up to \$50000. Anyone else is declined credit. This is clearly the most ``equal treatment'' possible, in the spirit of non-discrimination ``on the  basis of race, color, religion, national origin, sex, marital status, age, receipt of public assistance'', as mandated by the Equal Credit Opportunity Act.
At the same time, if one subgroup (defined by whichever protected attribute, e.g., race or the  receipt of public assistance) is having a lower-than-average income, its default rate on the \$50000 loan may be higher than that of the other subgroups. Over time, the subgroup with lower-than-average income will be regularly declined credit as a result of these defaults, in violation of the ``equal impact''.
On the other hand, if the credit limit is, e.g., set at three times the annual salary, the subgroup with lower-than-average income will be offered lower credit limits, in violation of the ``equal treatment''. The differentiated credit limits may make it possible for the same subgroup to repay the loans successfully, though, to develop a credit history, and eventually lead to a positive and ``equal impact''.\footnote{
While the Equal Credit Opportunity Act mandates that one must accurately describe the factors actually scored by a creditor, it does not suggest which of the above is preferable.
Specifically, it says ``if creditors know they must explain their decisions ... they [will] effectively be discouraged from discriminatory practices''.
}
See the penultimate section of this paper for further details of the application. 

Our original contribution then stems from the reinterpreting of the meaning of equal treatment and equal impact within a closed-loop view of the AI system.
There, an AI system produces information, which is communicated to the users, who respond to the information. The aggregate actions of the users are observed and serve as an input to further uses of the AI system.
Equal treatment concerns a single run of this closed-loop, while equal impact concerns long-run properties of this closed-loop.

The closed-loop view of the AI system addresses several important shortcomings of the presently proposed systems:
\begin{itemize}
    \item it very clearly distinguishes equal impact from equal treatment;
    \item it allows for a stochastic response of the users to the information produced by the AI system, rather than assuming it is deterministic;
    \item it explicitly models the ``concept drift'' and retraining of the AI system over time, inherent in practical AI systems, but ignored by most analyses of AI systems.
\end{itemize}
In terms of technical results, we formalise the notions above, present one condition that is necessary for the equal impact of an AI system, and illustrate the notions on a credit-risk use case.  

% https://pkghosh.wordpress.com/2020/12/24/concept-drift-detection-techniques-with-python-implementation-for-supervised-machine-learning-models/
% https://www.crc.business-school.ed.ac.uk/sites/crc/files/2021-04/Scorecard-Modelling-Best-Practice-Forrest.pdf

\section{Related Work}

\subsection{Regulation of AI}

While there is a long history of research on the interface of AI and law \cite[e.g.]{bench2012history,narayanan2018translation,berente2021managing}, much recent interest \cite[e.g.]{smuha2021race,petit2021models} has been sparked by the plans to introduce AI regulation within the legal system.
By investigating the self-regulation of leading AI companies from both the USA and Europe, \cite{de2021companies} appeal for future practices and governmental regulation.
Arguably, the European Commission regulates AI already: Article 22.1 of the General Data Protection Regulation (GDPR) is sometimes interpreted as prohibiting fully automated decisions with legal effect or ``similarly significant effect''.
There is much discussion regarding the AI Act \cite{veale2021demystifying} and regulatory landscape \cite{bringas2022fairness,vokinger2021regulating} in the Europe Union, and the potential extensions of the regulatory framework in the USA \cite{chae2020us}.
The EU Artificial Intelligence Regulation Proposal, sugguests use of ``feedback loops'' that perform the detection of biased outputs and the repeated introduction of appropriate methods of bias mitigation. \footnote{Article 15 of this Proposal emphasises that ``High-risk AI systems that continue to learn after being placed on the market or put into service shall be developed in such a way to ensure that possibly biased outputs due to outputs used as an input for future operations (‘feedback loops’) are duly addressed with appropriate mitigation measures.''}

Within the recent discussions, a fair amount of attention focuses on the question of defining AI \cite{schuett2019legal} -- or whether one should like to regulate the use of any algorithm \cite{schuett2019defining,ellul2022should} -- and defining high-risk uses of AI. 
One would also like to distinguish \cite{smuha2021beyond} between the harm of the individual and the society.
Further, in high-risk applications of AI, the automated decision-making AI systems are bound to be fair while formalisation of fairness definitions has been a long-standing debate. From the prospectives of fair outcomes, group fairness, such as demographic parity \cite{calder2009experimental}, equal opportunity \cite{hardt2016equality}, requests people from protected groups to be given the same treatment as others, while individual fairness requests ``similar people to be treated similarly'' \cite{petersen2021post,dwork2012fairness}. On the other hand, casual fairness \cite{chiappa2019path,kusner2017counterfactual} asks for a fair decision process, such that protected attributes are not direct causes of decisions, or only through certain causal paths.
Some recent works have extended to defining fairness in specific contexts, using users' feedback \cite{wen2021algorithms,d2020fairness,awasthi2020beyond}.

In contrast, we distinguish between the treatment within a single interaction with the AI system and the impact of repeated interactions with the AI system. Further, we propose a closed-loop framework that repeatly increases fairness, using aggregated feedback or users' responses. 

%\cite{dobbe2021hard}
%\cite{petit2021models}

\subsection{Control Theory}

Our approach is rooted in the closed-loop view of feedback control, but with several important differences.

%\textcolor{red}{To begin, conventional control is usually focused on regulating a single system.}
%Conventional control has a focus on regulating a single system.
Classic control often focuses on regulating a single system.
The system achieves the required behaviour most efficiently given the restrictions imposed by the challenge and the available resources. Even in areas where large-scale coupled systems are studied, the behaviour of all system components is analyzed and developed. On the other hand, in artificial intelligence, it is not the behaviour of individual users that is of interest. Rather than that, the variable of interest is the aggregate impact of the acts of a large number of users.
%\textcolor{red}{Synchronizing behaviour is unneeded,} if not detrimental because it is not even essential for all users to act rationally.
Examples of this kind of analysis include demand management for shared resources such as water and electricity, and the provision of medical care. De-synchronization alleviates the supply strain, and collective effects quantify the supply's quality. On the other hand, limits on the needed level of service for persons vary according to the application area.

Second, classical control, in general, is concerned with the control of systems with fixed dimensions. On the other hand, artificial intelligence often regulates and affects the behaviour of large-scale populations. Even the system's dimensions may be unpredictable and variable in such settings, emphasizing the critical requirement for scale-free management of extremely large-scale systems. Except in the case of passive control design, scale-free control for large systems is a largely unexplored issue in the classical control field.

Thirdly, in classical control, the controlled system's mathematical description does not change in response to control signals. This underlying concept is challenging to realize in artificial intelligence. By and large, models can only approximate the dynamics of the actual systems. This is not an issue as long as there is an appreciation for the possibility of reality and model deviating from one other. However, models in artificial intelligence are not easily derived from first principles; instead, they are empirical, i.e., based on data gathered from measurements of existing processes. Additionally, controlled studies cannot gather empirical data across a variety of operating points but must be obtained directly from the system. 

An effort to enhance the processes above, for example, by sending information to the users involved, establishes a feedback loop that did not exist earlier. This change in the underlying process may invalidate the empirical model since there were no data available to represent the dynamic influence of such feedback during the model's development. Frequently, offered solutions ignore this feedback loop. This latter aspect necessitates a far more extensive examination of prediction and optimisation under feedback than has hitherto been the case. 
%Signalling to mitigate this feedback effect is a hot topic of discussion at the moment and represents an intriguing extension of classical pricing theory.

Fourth, data sets are often gathered in a closed-loop fashion like  Figure~\ref{fig:ch3-system}. That is, public data sets often contain information about decision-makers. Developing models of large-scale feedback systems is a crucial hurdle to development in applying certain control methods to artificial intelligence. In dealing with such impacts, artificial intelligence researchers may have a lot to learn from economic and control theory.

Finally, and perhaps most significantly, a fundamental distinction between classical control and our approach is the need to investigate the influence of control signals on the statistical features of the populations under control. Given that we are often dealing with service delivery, these statistical features should be stationary and predictable, necessitating ergodic control design. %We are particularly interested in establishing systems such as the emphasizing-sharing economy, in which economic contracts can be appropriately specified. 
%Our starting point is that many challenges in smart cities may be framed in terms of a competition between several users, such as people, automobiles, or robots, over a finite resource. The challenge of allocating this resource in a way that is not wasteful that maximizes the return on the resource's use for society, and that also provides a guaranteed level of service to each of the users competing for that resource raises a slew of issues that are best addressed using control theory. From a control engineer's viewpoint, this statement may be deconstructed into three goals, two of which are well-known in the field of control and one of which is a relatively new concept.

\subsection{Control of Multi-user Dynamical Systems}

Perhaps the closest to our work within control theory are multi-user dynamical systems over networks.
There, the principal concern is the design of distributed protocols that provide consensus or synchronisation of states of all users \cite{Blondel2005,Nedic2009}. (The states might indicate vehicle directions or locations, estimations of sensor readings in a sensor network, oscillation frequencies, and each user's trust opinion, among other things.) To achieve synchronised behaviour in multi-user systems, all systems must agree on the values of these quantities.

Studying their interactions and collective behaviours under the effect of the information flow permitted by the communication network is critical for networked cooperative dynamical systems.
This communication network may be seen as a graph with directed edges or connections corresponding to the information travelling between the systems. The systems are portrayed as nodes on the graph and are sometimes referred to as users. In communication networks, information flows exclusively between the graph's close neighbours. However, if a network is linked, this locally sent information eventually reaches every user in the graph.

In cooperative control systems based on graphs, there are fascinating interactions between the dynamics of the individual users and the communication graph's topology.
The graph topology may severely constrain the performance of the users' control rules. To be precise, in cooperative control on graphs, all control protocols must be distributed so that each user's control rule is limited to knowledge about its near neighbours in the network topology. If sufficient attention is not taken while constructing the local user control rules, the dynamics of the individual users may be stable, but the graph's networked systems may display undesired behaviours. Due to the communication constraints imposed by graph topologies, complex and fascinating behaviours are seen in multi-user systems on graphs that are not found in single-user, centralised, or decentralised feedback control systems.

The ideas of distributed cooperative control are used in \cite{Lewis2013} to construct optimal and adaptive control systems for multi-user dynamics on graphs. The requirement complicates these designs that all control and parameter tweaking methods must be dispersed in the network to rely on just their near neighbours. 

\cite{Lewis2013} analysed discrete-time systems and demonstrate that an additional condition between the local user dynamics and the graph topology must be met to ensure global synchronization when the local optimum design is used. Global optimization of collective group movements is more challenging than locally optimizing each user's motion. A typical issue in optimum decentralized control is that global optimization problems often demand knowledge from all users, which distributed controllers cannot access since they can only utilize information from closest neighbours. Further, they demonstrate, globally optimum distributed form controls may not exist on a particular graph. To achieve globally optimum performance when employing distributed protocols that rely only on local user information in the graph, the global performance index must be chosen to depend on graph features, notably the graph Laplacian matrix. They also establish distinct global optimality for which distributed control solutions are always possible on sufficiently linked networks. There, they examine multi-user graphical games and demonstrate that a Nash equilibrium results when each user optimizes its local performance index. For more results on these direction we refer \cite{Shamma2008, Wang2017, Wang2021, Yu2017, Chen2019}.

%Some specific applications in ITS are monitoring and controlling transportation flows \cite{Chen2010}, CO$_2$ emissions \cite{Arieh2013}, assigning parking spaces \cite{Arieh2014}, and utilising vehicle-to-grid (V2G) energy supply \cite{Shaukat2018,Pillai2010}.
%In the context of smart grids, problems related to real-time electricity pricing \cite{Mohsenian2010}, real-time demand response \cite{Conejo2010,Callaway2010}, and real-time interruptible loads management \cite{Herriges1988,Alagoz2013,Salsbury2013}.
%Moreover, V2G-related issues have also attracted interest over the last few years. For these -- and many other -- applications, \cite{Fioravanti2019} present both conditions ensuring and ruling out ergodicity in a certain closed-loop sense. 

\section{A Closed-Loop View of AI Systems}

Let us consider a closed-loop model based on the following constraints:
\begin{itemize}
%\item Aiming to allocate a limited resource among several users periodically based on information supplied by a central authority on the resource.
\item Users get information from the AI System, but are not required to take action based on the AI System's outputs. It will be convenient to encode user's reaction to the output probabilistically.
\item The AI System does not necessarily monitor individual user's actions (``profiling''), but rather some aggregate or otherwise filtered version. 
\item The users do not communicate with one another, or only in response to information broadcast by the central authority.
\end{itemize}

%\subsection{Mathematical Description of the Closed-loop Feedback model}\label{subsec:ch3-math-des-mod}
\begin{figure}[t!]
\centering
\includegraphics[width=\columnwidth]{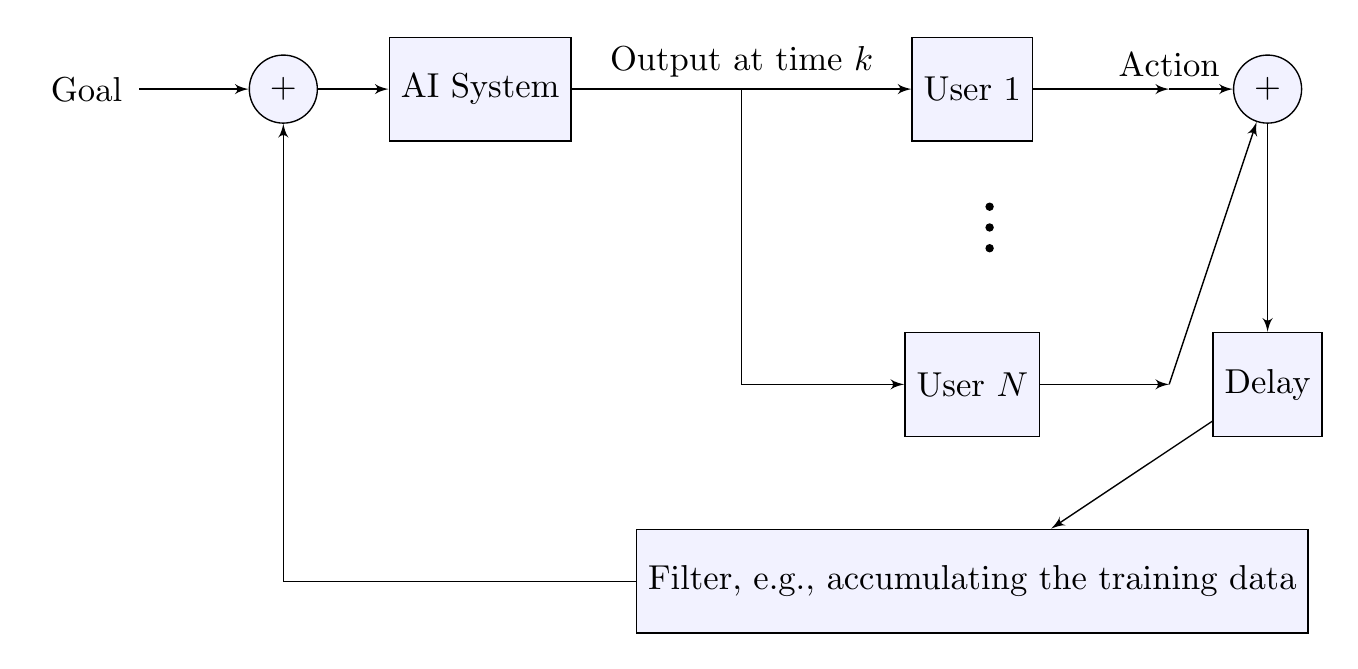}
\caption{A closed-loop model of an AI system and its interactions with the users: the AI system provides some outputs, e.g., scorecards in credit scoring, matches in a matching market, or suggestions in a decision-support system. Users observe the outputs and take action in response. With some delay, their actions in response to the outputs are utilized in retraining the AI System.}
\label{fig:ch3-system}
\end{figure}

Ultimately, the repeated uses of an AI system can be seen as the closed-loop of Figure \ref{fig:ch3-system}.
The AI System produces some outputs $\pi(k)$ at time $k$, e.g., lending decisions in financial services, matches in a two-sided market, or suggestions in a decision-support system. 
The output is taken up by $N$ users of the system, who 
have some states $x_i$, $i\in[N]$ internal to them, where $[N]=1,\ldots,N$.
%In particular, $x_i\left(k\right)$ is a random variable. 

The users take some action, which can be modelled as a probability function of the output and the private state, over the certain user-specific sets of actions.
%This can be modelled by user-specific and output-specific probability distributions over the certain user-specific sets of actions $\mathbb{A}_i =\{a_1,\ldots,a_L \}\subset \mathbb R^{n_i}$, where $\mathbb R^{n_i}$ can be seen as the space of $i^{\textit{th}}$ user's  private state space $x_i$. 
The action $y_i\left(k\right)$ of user $i$ at time $k$ is then a random variable. 
In the remainder, we will assume $y_i\left(k\right)$ are scalars, but generalisations are easy to obtain.
%The randomness can be a result of the inherent randomness in the reaction of user $i$ to the control signal $\pi(k)$, or the response to a control signal that is intentionally randomized \cite{Arieh2013,Arieh2014,Marecek2015}. 
The aggregate of the actions  $y(k)=\sum\limits_{i = 1}^{N} y_i\left(k\right)$ at time $k$ is then also a random variable. 
The AI System may not have access to either $x_i\left(k\right)$, $y_i\left(k\right)$, but perhaps only  $y(k)$ or some filtered version. 
The filter may accumulate the data, for instance, before filtering out anomalies. 

%we assume thatincluding t the controller has its private state $x_c\left(k\right) \in \mathbb R^{n_c}$. The controller aims to regulate the system by providing a signal $\pi(k) \in \Pi \subseteq\mathbb R$ at time $k$; here, $\pi(k)$ denotes the set of admissible broadcast control signals. In the simpler static case, the signal $\pi(k)$ is a function of an error signal $e(k)$ and the controller state $x_c(k)$, with range $\pi(k)$.

\section{Equal Treatment}\label{sec:eqtreatment}

Equal treatment very clearly examines the AI system's treatment of its users and the influence on the microscopic qualities over the short run. 

\begin{definition}[Equal Treatment]
For each user $i$, we require that
\begin{itemize}
 \item [i)] the system provides the same information $\pi(k)$ to all users $i$,
 \item [ii)] that there exists a constant $\overline{r}$ such that
\begin{equation}\label{eq:ch3-treatmentt}
y_i(j) = \overline{r},
\end{equation}
where this constant is independent of initial conditions.
\end{itemize}
\end{definition}

\begin{definition}[Equal Treatment Conditioned on Non-Protected Attributes]
For each user $i$ within a class that is defined by non-protected attributes, we require that
\begin{itemize}
    \item [i)] the system provides the same information $\pi(k)$ to all users within the class;
    \item [ii)] that there exists a constant $\overline{r}$ such that
\begin{equation}\label{eq:ch3-treatment}
 y_i(j) = \overline{r},
\end{equation}
where this constant is independent of initial conditions. 
\end{itemize}
\end{definition}
Notice that there is a sufficiently large overlap of the classes that are defined by non-protected attributes such that the definition reduces to the unconditional equal treatment.

\section{Equal Impact}\label{sec:eqimpact}

Equal impact very clearly examines the AI system's influence on the user population's microscopic qualities over the long run. One may desire, for example, that each user obtains a fair portion of the resource on average over time, or, at a far more fundamental level, that the average allocation of the resource to each user over time is a stable number that is predictable and independent of beginning circumstances. 

To model equal impact, we construct requirements that ensure ergodicity: the presence of a single invariant measure to which the system is statistically drawn regardless of the starting circumstances.

\begin{definition}[Equal Impact]
For each user $i$, we require that
\begin{itemize}
    \item [i)] there exists a constant $\overline{r}_i$ such that
\begin{equation}\label{eq:ch3-pred}
\lim_{k\to \infty} \frac{1}{k+1} \sum_{j=0}^k y_i(j) = \overline{r}_i,
\end{equation}
where this latter limit is independent of initial conditions;
\item [ii)] all the $\overline{r}_i$ coincide.
\end{itemize}
\end{definition}
\begin{definition}[Equal Impact Conditioned on Non-Protected Attributes]
For each user $i$ within a class that is defined by non-protected attributes,
there exists a constant $\overline{r}_i$ such that
\begin{equation}\label{eq:ch3-predi}
\lim_{k\to \infty} \frac{1}{k+1} \sum_{j=0}^k y_i(j) = \overline{r}_i,
\end{equation}
where this latter limit is independent of the initial conditions.
Furthermore, we require that all the $\overline{r}_i$ coincide.
\end{definition}
 
\section{Guarantee Properties}

Proving that there is a unique invariant measure is not necessarily an easy undertaking. Even well-known AI systems do not always result in feedback systems that exhibit equal impact.

Under the assumptions of continuity of the closed-loop model, the work on \emph{iterated function systems} \cite{Elton1987,Barnsley1989,Diaconis1999}, which are a class of stochastic dynamical systems arising from the multi-user interactions, makes it possible to obtain strong stability guarantees for such stochastic systems under the assumptions of continuity of the closed-loop model.
The following are shown in the work \cite{Fioravanti2019}:
\begin{itemize}
\item Even if regulation is accomplished by controlling the behaviour of ensembles of users, feedback control with integral action has the potential to disrupt the closed-loop system's ergodic features. This discovery is significant because ergodic behaviour is necessary for supporting economic contracts and ensuring the existence of attributes such as fairness. Thus, from a practical standpoint, the finding is one of the system's critical features and is not only theoretically interesting.

\item A few particular instances are given to demonstrate the loss of ergodicity in seemingly innocuous situations.

\item For particular population types and filters, stable control action always results in ergodic behaviour. It was particularly shown for linear and non-linear systems with both real-valued and finite-set actions.

\item Finally, a minor contribution was made to demonstrate how the results from the study of iterated function systems might be used in designing controllers for specific types of dynamic systems.
\end{itemize}
In this paper, we have to relax the continuity assumptions, however. Indeed, the classification problems involve discrete sets such as the ``credit denied'' or ``credit approved'', which cannot be easily modelled with continuous fuctions. So in this case, stochastic, user-specific response to the feedback signal $\pi(k)\in \Pi$ can be modelled
by user-specific and signal-specific probability distributions over the certain user-specific sets of actions
\begin{align}\label{eq:action}
\mathbb{A}_i =\{a_1,\ldots,a_L \}\subset \mathbb R^{n_i},   
\end{align}
where $\mathbb R^{n_i}$ can be seen as the space of $i^{\text{th}}$ user's  private state space $x_i$. Assume that the set of possible resource demands of user $i$ is
$\mathbb{D}_i$, where in the case that $\mathbb{D}_i$ is finite we denote
\begin{equation}\label{eq:demands}
\mathbb{D}_i := \{ d_{i,1},d_{i,2}, \ldots, d_{i,m_i}\}.
\end{equation}
In the general case, we assume there are $\tau_i \in \mathbb N$ state transition maps
\begin{align*}
w_{ij}: \mathbb R^{n_i} \to \mathbb R^{n_i}, j=1,\ldots,\tau_i
\end{align*}
for user $i$ and  output maps
\begin{align*}
w'_{i\ell}: \mathbb R^{n_i} \to\mathbb{D}_i, \ell= 1,\ldots,\kappa_i, \kappa_i \in \mathbb N,
\end{align*}
for each user $i$.
The evolution of the states and the corresponding demands then satisfy:
\begin{subequations}
\begin{align}
x_i\left(k+1\right) &= w_{ij}\left(x_i(k)\right) \;\vert\; j = 1, \ldots, \tau_i,\label{eq:state} \\
y_i(k) &= w'_{i\ell}\left(x_i(k)\right) \;\vert\; \ell = 1, \ldots, \kappa_i,\label{eq:output}
\end{align}
\end{subequations}
where the choice of user $i$'s response at time $k$ is governed by probability functions
\begin{subequations}
\begin{align}
p_{ij} : \Pi \to [0,1], j=1,\ldots,\tau_i\label{eq:prob-1}\\
p'_{i\ell} : \Pi \to [0,1], \ell=1,\ldots,\kappa_i,\label{eq:prob-2}
\end{align}
\end{subequations}

respectively. Specifically, for each user $i$, for all $k\in\mathbb N$ and for all signal $\pi$  we have that: 
\begin{subequations}
\begin{align}
&\mathbb{P}\big(x_i\left(k+1\right) = w_{ij}\left(x_i(k)\right) \big) = p_{ij}\left(\pi(k)\right),\label{eq:prob-3}\\
&\mathbb{P}\big(y_i(k) = w'_{i\ell}\left(x_i(k)\right)\big) = p'_{i\ell}\left(\pi(k)\right),\label{eq:prob-4}\\
&\sum\limits_{j=1}^{\tau_i} p_{ij}\left(\pi\right) = \sum\limits_{\ell=1}^{\kappa_i} p'_{i\ell}\left(\pi\right) = 1.\label{eq:prob-5}
\end{align}
\end{subequations}

%\begin{theorem}
%Consider the feedback system depicted in
%Figure 1. Assume that the output of the AI system at time $k$ is takes values from a discrete set, and actions of the users takes values from a discrete set \eqref{eq:demands}, following
%by the non-linear stochastic difference equations \eqref{eq:state}--\eqref{eq:output}.
%Assume we have Dini continuous probability functions
%$p_{ij}, p'_{il}$ so that the probabilistic laws %\eqref{eq:prob-1}--\eqref{eq:prob-5} are satisfied. Assume furthermore that there exists scalars $\delta,\delta' > 0$ such
%that $p_{ij}(\pi) \ge \delta$ and $p'_{il}(\pi) \ge \delta'$  for all $(i, j)$ and all $\pi$. Then, for AI systems that satisfy one of the following properties:
%\begin{enumerate}
%\item AI system is linear and stable \label{linear}
%\item AI system is input-to-state stable \label{iss}
%\item AI system has stochastic Lyapunov function ... 
%\label{sL}
%\end{enumerate}
%the following holds:
%If the graph $G = (X, E)$ is strongly connected, then there
%exists an invariant measure for the feedback loop. If in
%addition, the adjacency matrix of the graph is primitive,
%then the invariant measure is attractive and the system
%is uniquely ergodic.
%\end{theorem}

%\begin{proof}
Then, one can prove that when the graph $G = (X, E)$ is strongly connected, there
exists an invariant measure for the feedback loop. If in
addition, the adjacency matrix of the graph is primitive,
then the invariant measure is attractive and the system
is uniquely ergodic.

For linear systems, this is a direct consequence of (Werner, 2004) and the observation that the necessary contractivity properties follow from the internal asymptotic stability of controller and filter.
For non-linear systems, similar results can be obtained using \cite[Theorem 2]{Marecek2021}. See also \cite{ghosh2021ergodic} and the Supplementary information. 
%\end{proof}

\section{Numerical Illustrations}

Credit scoring refers to the process of lenders, usually financial institutions, measuring the creditworthiness of a person or a small business, usually derived from its historical default.
In USA, Equal Credit Opportunity Act (ECOA) and the part of the law that defines its authority and scope, known as Regulation B, require statements of specific reasons for adverse credit decisions, where it would be difficult, yet impossible to comply if complex algorithms or ``black-box'' models are used. %\footnote{\url{https://www.consumerfinance.gov/compliance/circulars/circular-2022-03-adverse-action-notification-requirements-in-connection-with-credit-decisions-based-on-complex-algorithms/}}.
Instead, scorecards are commonly adopted in practice, due to their good explainability, while alternatively, counterfactual explanations  \cite{dutta2022robust,verma2020counterfactual} work as an explainer of ``black-box'' models to guide an applicant on the easiest improvement that could change the model outcome.
Table~\ref{tab:scorecard} displays a simple scorecard.
%coef: [-9.99999999633862, -9.611548838262998, 0.38845005191903337, 10.388449689990752, 20.32117686758051, 22.522215350216236, 22.880189946417655]

%-9.999999995835388, -9.55797901537736, 0.4420200068960295, 10.442019588691158, 20.39431529237211, 22.811560117289524

\begin{figure}[t]
    \centering
    \includegraphics[width=0.35\textwidth]{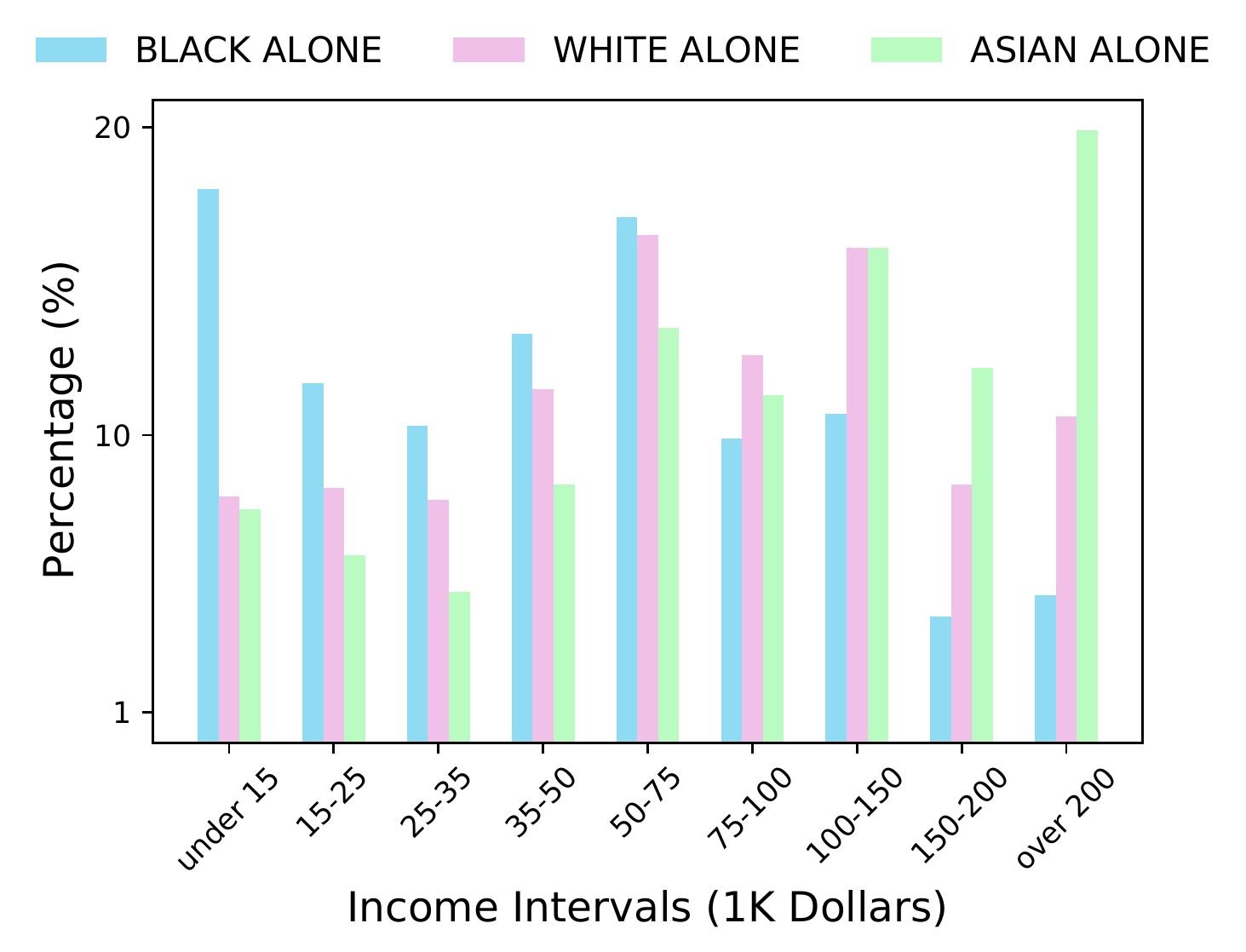}
    \caption{The 2020 annual income distribution of ``BLACK ALONE'', ``WHITE ALONE'' and ``ASIAN ALONE'' households in USA, with three races distinguished by colours. Data are sourced from Table A-2 of the Current Population Survey (CPS) of US Census Bureau.}
    \label{fig:percentage}
\end{figure}

%-8.16851986714064, -6.675117259843017e-11, 5.768473174850463
%-8.220528806317214, -1.8539532165832718e-10, 5.694111460573914
\begin{table}[ht]
\centering 
\begin{tabular}{c c c c}
\hline\hline
Factor & Code & Description & Score \\[0.2ex]
\hline 
History & - & $\times$ Average Default Rate & -8.17 \\
\hline 
\multirow{2}{*}{Income} & 0 & $\leq \$ 15\textrm{K} $ & 0 \\ 
& 1 & $> \$ 15\textrm{K} $ & +5.77 \\[0.1ex]
\hline\hline
\end{tabular}
\caption{A simple scorecard for existing users. For example, a user with annual income $\$ 50\textrm{K}$ and an average default rate $0.1$ would be given a score of $ -8.17\times 0.1 + 5.77=4.953$.}
\label{tab:scorecard}
\end{table}
Although Table~\ref{tab:scorecard} might seem fair at first sight, income is a factor closely related to protected attributes, e.g., race. Figure~\ref{fig:percentage} displays the 2020 annual income distribution of households by race, including ``BLACK ALONE'' (blue), ``WHITE ALONE'' (pink) and ``ASIAN ALONE'' (green), in the USA, sourced from \textit{Table A-2. Households by Total Money Income, Race, and Hispanic Origin of Householder: 1967 to 2020} (Table A-2), from US Census Bureau \footnote{See \url{https://www.census.gov/data/tables/2021/demo/income-poverty/p60-273.html}}. The green bar on the index ``over 200'' implied that a larger share (almost 20\%) of ``ASIAN ALONE'' households makes more than $\$ 200\textrm{K}$ in 2020. 
On the other hand, the income of most ``BLACK ALONE'' households is less than $\$ 75\textrm{K}$. 
This figure casts doubt on the equal treatment using the scorecard in Table~\ref{tab:scorecard}, because races with generally lower incomes would receive a lower credit score.
If a lender tries to maintain similar credit distributions across different races, the results may not be as expected in the long run, as low-income households might end up defaulting or even not be able to apply for another mortgage ever after, thus hurting their long-term credit history.

Our notion of equal impact in the context of credit scoring would equalise the long-term average default rate across races or across individuals, such that low-income households can keep better credit history.
%, and hence narrow down the gap of long-term results.
%against whom adverse action is taken more frequently.
Recall Figure~\ref{fig:ch3-system} from the perspective of credit scoring. 
Given the goal of equal impact, at each time step, the income 
$z_i(k)$ is internal to the user (user), while her income code $\mathbbm{1}_{\geq 15}z_i(k)$ is visible to a lender, where $\mathbbm{1}_{\geq 15}$ is an indicator function that maps the input to one if $\geq 15$ is satisfied and all other values to zero.
The lender would use the AI system, i.e., logistic regression in our case, to build a scorecard and reveal a credit decision $\pi(k,i)$ (e.g., approval or denial of a mortgage transaction) to user $i$ at time $k$. 
Note that the scorecard only gives a credit score, but, based on a cut-off score, the lender is able to reach a credit decision.
%$\pi(k)$ at time $k$ and reveal it to all users (users). 
%Following the credit decision $\pi(k,i)$, user $i$ would be assigned a credit score, and then given a credit decision (e.g., approval or denial of a mortgage transaction), based on a cut-off score.
Confidential to cilent $i$, her state $x_i(k)$ at time $k$ is determined by her income and, in turn, influences the repayment action.
%For simplexity, we assume the first element of its state (i.e., credit decision) is binary $x_i^1(k)\in\{0,1\}$, where $x_i^1(k)=1$ denotes that user $i$ is offered a 3.5-times-income mortgage.
Its repayment action $y_i(k)\in\{0,1\}$ is modelled as a Gaussian conditional independence model \cite{tang2021quantum,leitao2020model,rutkowski2015regulatory}. % TODO: More refs needed other than egger2020credit
Afterwards, the filter calculates the average default rates of each user, using historical repayment actions $y_i(k)$ for $i\in 0,\ldots,k$.
The average default rates, along with the income code of users, would be used as training data for the AI system, and further, new credit decisions $\pi(k+1,i),i\in[N]$ are made again using logistic regression. 
%Afterwards, the income $z_i(k)$ (i.e., the second element of the state $x_i(k)$) and historical repayment actions $y_i(k)$ for $i\in 0,\ldots,k$ would be used as training data for the AI system, and a new scorecard $\pi(k+1)$ is built again using logistic regression. 

For the numerical experiments, we use the real-world data from Table A-2, which gives the number of households and income distribution by year and race. 
We consider a period from 2002 to 2020, with a year being a time step, because in 2002 the Annual Social and Economic Supplement (ASEC) of the Current Population Survey (CPS) started to allow households to report their race from more diverse options.
Let $\mathcal{S}$ be a set that includes 3 races: ``BLACK ALONE'', ``WHITE ALONE'' and ``ASIAN ALONE''.
In the beginning of 2002 (time $0$), we generate $N=1000$ users (households), whose races are sampled from $\mathcal{S}$ with a distribution of $[0.1235,0.8406,0.0359]$. 
Notice that the distribution is the ratio of the number of households of the three races in 2002 in Table A-2.
The generated user set is then divided into 3 subsets according to race, denoted by $\mathcal{N}_s$, for $s\in\mathcal{S}$.
Further, following the income distribution of the year $2002+k$ and race $s$, we sample the income $z_i(k)$ of user $i\in\mathcal{N}_s$ at time $k$.

For simplicity, let $\pi_i(k,i)=1$ denote that user $i$ is offered a 3.5-times-income mortgage at time $k$.
Assuming that the annual mortgage rate and the basic living cost are 2.16\% per annum and $\$ 10\textrm{K}$, we use the Gaussian conditional independence model \cite{rutkowski2015regulatory} to generate the repayment actions.
Suppose that the state $x_i(k)$ measures the portion of income left after deduction of living cost and mortgage interest: %\eqref{eq:repay-latent}:
\begin{equation}
x_i(k)=\frac{z_i(k)-10-3.5\times 2.16\%\times z_i(k)}{z_i(k)}.
\label{eq:repay-latent}
\end{equation}
%where $z_i(k)$ denotes the second element of the state $x_i(k)$ (i.e., income). 
The binary repayment action $y_i(k)$ (1 for repaid) is defined by \eqref{eq:repay-generation}.
%Furthermore, the repayment action $y_i(k)\in\{0,1\}$ is modelled as a Gaussian conditional independence model \cite{egger2020credit}, such that
\begin{equation}
\begin{cases}
y_i(k) =0 & 
\begin{aligned}
&\text{for } x_i(k) \leq 0\\
& \; \text{ or }\pi(k,i)=0,
\end{aligned}\\
y_i(k) \sim \textrm{Bernoulli}\left( F\left(5\times x_i(k)\right)\right) & \text{otherwise, } 
\end{cases}
\label{eq:repay-generation}
\end{equation}
where user $i$ would not make a repayment if no mortgage is offered or if her income cannot cover the basic living cost plus mortgage interest. Otherwise, the repayment action follows a Bernoulli distribution with $\textrm{Pr}(y_i(k)=1)=F(5\times x_i(k))$, where $F(\cdot)$ is the cumulative distribution function of the standard normal distribution. 

Furthermore, 
we define default as a mortgage offered but not repaid, i.e., $y_i(k)=0|\;\pi(k,i)=1$.
We introduce the average default rate $\textrm{ADR}_i(k)$ for user $i$ and the race-wise version $\textrm{ADR}_{s}(k)$ for race $s$ at time $k$, as defined in \eqref{eq:ADR}:
\begin{equation}
\begin{split}
\textrm{ADR}_i(k):=&\textrm{Pr}(y_i(k)=0| \pi(k,i)=1)\\
=&1-\sum_{j=0}^{k}\frac{y_i(j)}{\pi(k,i)}\\
%\frac{1}{k+1}\sum_{j=0}^k y_i(j),\qquad\qquad\quad\forall i\in [N],\\
\textrm{ADR}_{s}(k):=&\textrm{Pr}(y_i(k)=0| \pi(k,i)=1, i\in \mathcal{N}_s)\\
=&1-\frac{1}{|\mathcal{N}_s|}\sum_{i\in \mathcal{N}_s}\sum_{j=0}^{k}\frac{y_i(j)}{\pi(k,i)},
%\frac{1}{(k+1)|\mathcal{N}_s|}\sum_{j=0}^k \sum_{i\in \mathcal{N}_s} y_i(j), \quad\forall s\in\mathcal{S},
\end{split}
\label{eq:ADR}
\end{equation}
where $|\mathcal{N}_s|$ denotes the number of users of race $s$. 
With the goal of equal impact, we wish to equalise the outcome of credit scoring among individuals in the long run, such that 
\begin{equation}
%\begin{split}
\lim_{k\to\infty}\textrm{ADR}_i(k)=\overline{r}_i,\quad
\lim_{k\to\infty}\textrm{ADR}_s(k)=\overline{r}_s,
%\end{split}
\label{eq:goal-ADR}
\end{equation}
and that all $\overline{r}_i$ coincide and all $\overline{r}_s$ coincide.

\begin{figure*}[!htb]
    \centering
    \begin{minipage}{.33\textwidth}
        \centering
       \includegraphics[width=\columnwidth]{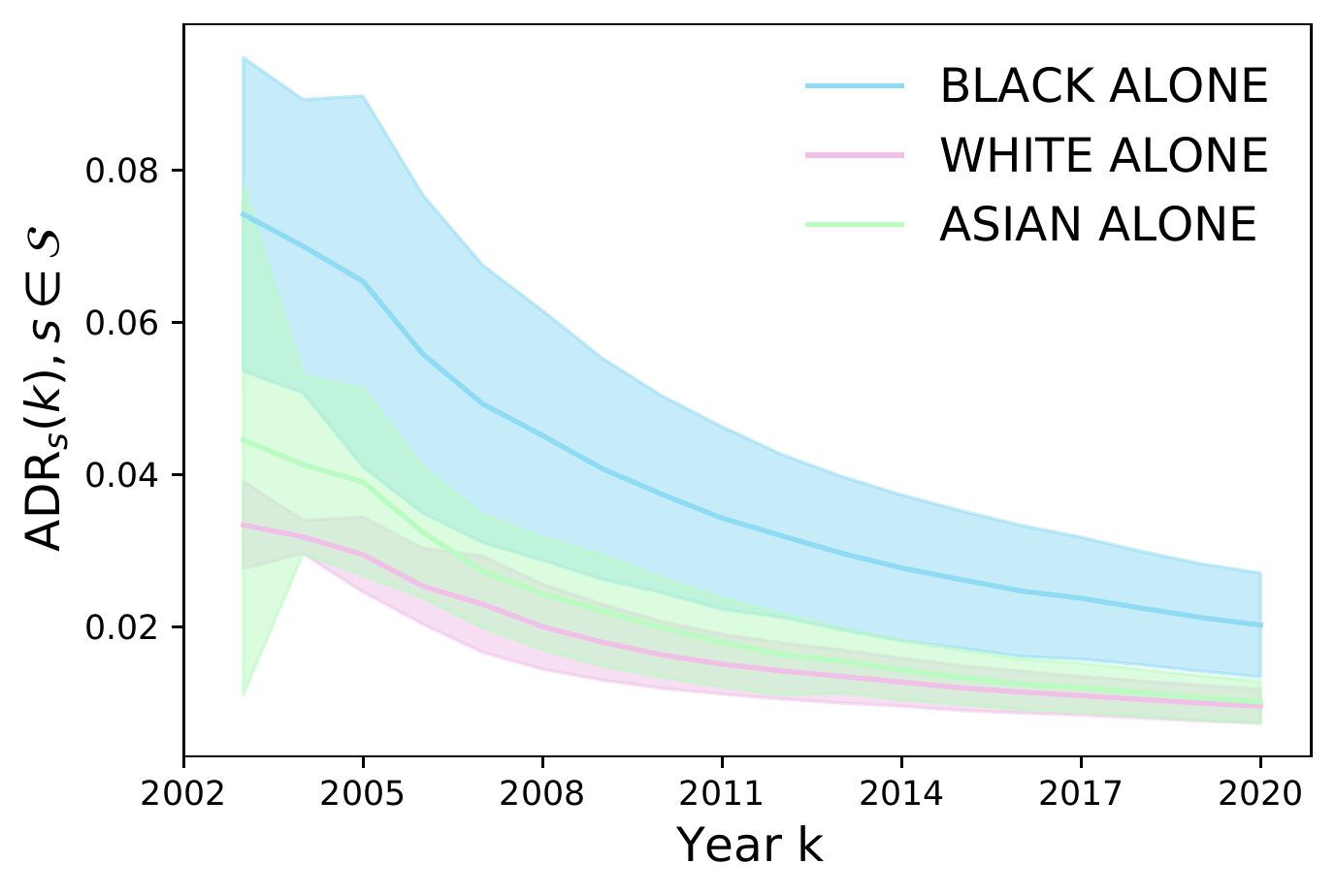}
    \caption{Solid curves depict the mean value of time series $\{\textrm{ADR}_s(k)\}_{k\in[N]}$, across five trials, with race information distinguished by colour. Error shades display mean $\pm $ one standard deviation.}
    \label{fig:group_perf}
    \end{minipage}%
    \hfill
    \begin{minipage}{0.33\textwidth}
        \centering
        \includegraphics[width=\columnwidth]{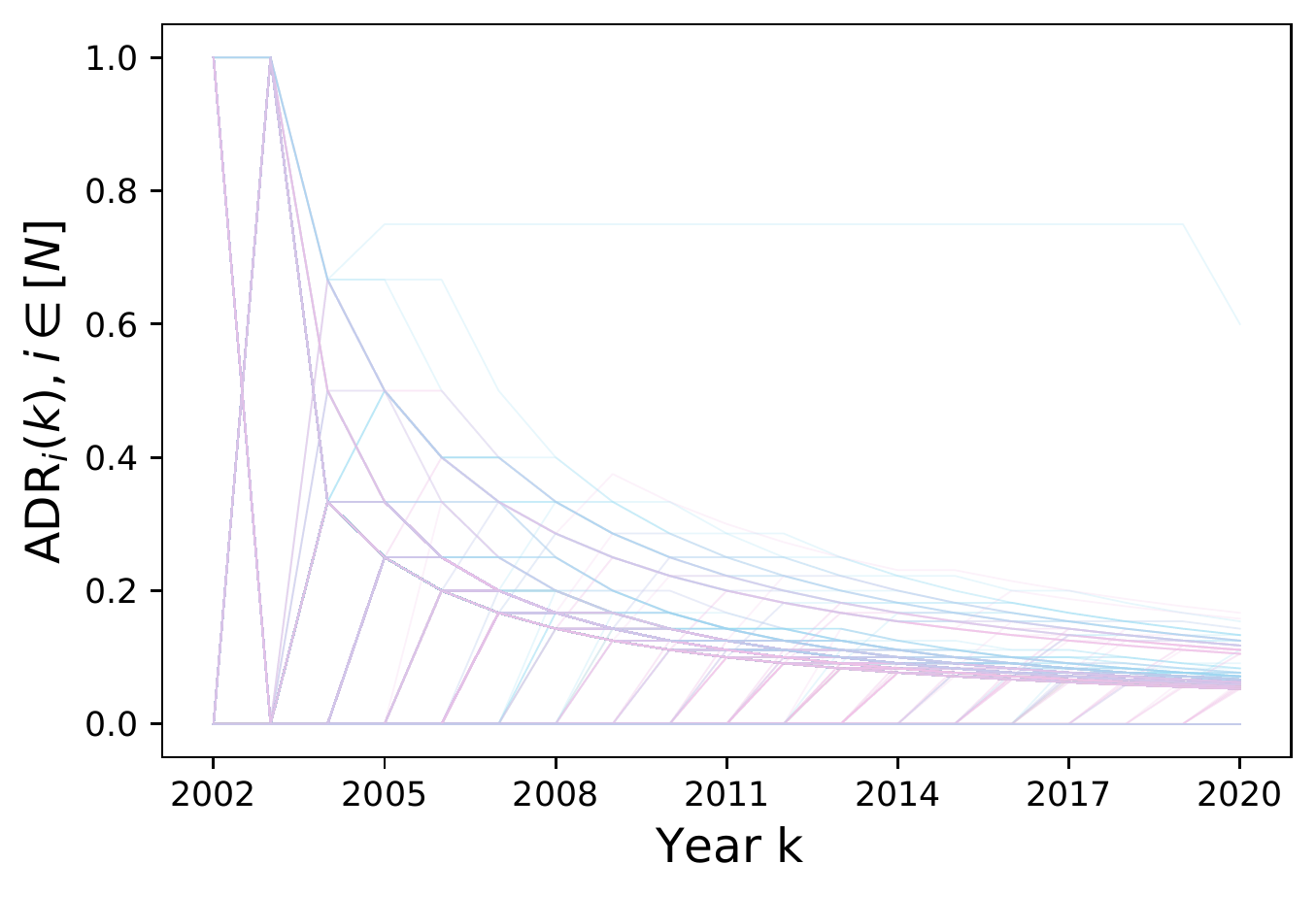}
    \caption{The time series $\{\textrm{ADR}_i(k)\}_{k\in[N]}$ for all users from five trials ($5\times 1000$ curves), with their race information distinguished by colour. \newline}
    \label{fig:indiv_perf_ts}
    \end{minipage}
    \hfill
\begin{minipage}{0.31\textwidth}
    \centering
 \includegraphics[width=\columnwidth]{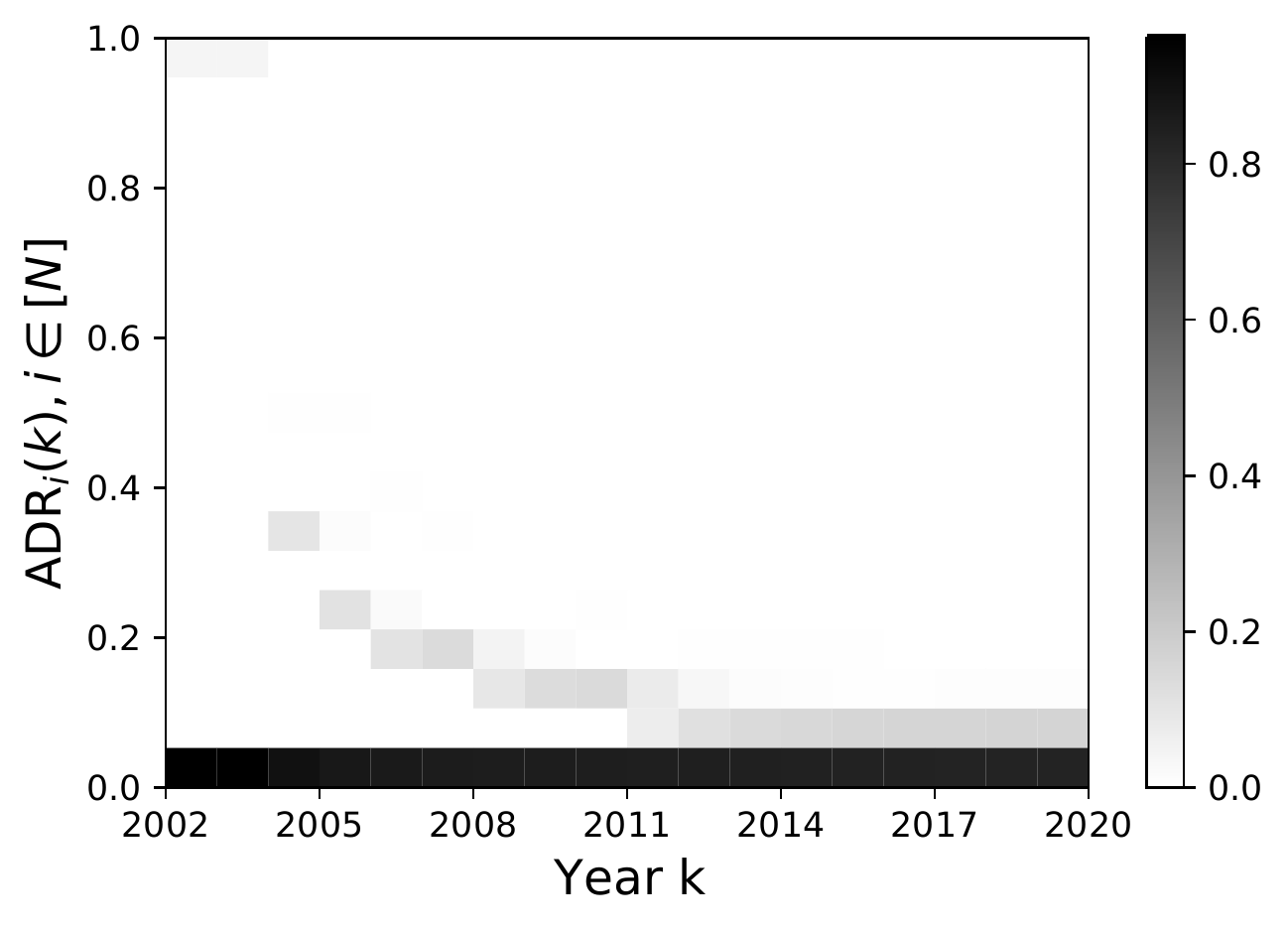}
\caption{The density of ADR$_i(k)$ at different time steps, with the race information ignored. Darker colours denote higher density.}
\label{fig:indiv_perf_heatmap}
    \end{minipage}
\end{figure*}

% \begin{figure}[!htb]
% \centering
% \includegraphics[width=0.4\textwidth]{indiv_perf_heatmap.pdf}
% \caption{The density of ADR$_i(k)$ at different time steps, with the race information ignored. Darker colours denote higher density.}
% \label{fig:indiv_perf_heatmap}
% \end{figure}

For the year of 2002-2003 (time 0 \& 1), no scorecard is used and we assume all users are given the approval of the mortgage, e.g., $\pi(k,i):=1$, for $i\in [N]$ and $k=\{0,1\}$.
Thus, we obtain the initilisation of average default rates, i.e., $\textrm{ADR}_i(0),\textrm{ADR}_{s}(0)$ and $\textrm{ADR}_i(1),\textrm{ADR}_{s}(1)$.
Afterwards, for time $k\geq 2$, a scorecard is built, whose parameters are trained from a logistic model, with independent variables being $\mathbbm{1}_{\geq 15} z_i(k)$, ADR$_i(k-1)$ and the dependent variable being $\ln{\frac{y_i(k)}{1-y_i(k)}}$. 
%Note that $\mathbbm{1}_{\geq 15}$ is an indicator function that maps the input to one if $\geq 15$ is satisfied and all other values to zero.
%\mathbbm{1}_{\geq 10}(z_i(k-1))$ and ADR$_i(k-2)$ for $i\in[N]$ for all users are used to train a logistic regression model for estimating the probability of the event $y_i(k-1)=1$. 
Although, the scorecard $\pi(k)$ can vary in time steps, we use the same cut-off score 0.4 to decide each user's credit decision (0 for denial and 1 for approval). %Let $\pi(k,i)$ be user $i$'s credit score at time $k$, according to the scorecard $\pi(k)$ and the information of $\mathbbm{1}_{\geq 15}(z_i(k))$, ADR$_i(k-1)$.
Using our notation, the example of Table~\ref{tab:scorecard} would be rewritten as
\begin{equation*}
\begin{split}
&-8.17\times \textrm{ADR}_i(k-1)+5.77\times \mathbbm{1}_{\geq 15} z_i(k)
= 4.953 \\ &
\qquad\quad\Rightarrow\quad 4.953 > 0.4 
\quad\Rightarrow\quad \pi(k,i)=1.
\end{split}
\end{equation*}

We define a trial as the simulation of generating 1000 users ($N=1000$) and repeating the closed-loop for the period 2002-2020.
In our numerical experiments, five trials are conducted, with each trial using a new batch of 1000 users.
%In a single trial, the sequence of $\{ \textrm{ADR}_i(k)\}_{k\in[N]}$ for a user $i$ could be seen as a time series. So did the sequence of $\{\textrm{ADR}_s(k)\}_{k\in[N]}$ for a race $s$.
For consistency with Figure~\ref{fig:percentage}, the races ``BLACK ALONE'', ``WHITE ALONE'', and ``ASIAN ALONE'' are represented by blue, pink, and green colours, respectively.

In Figure~\ref{fig:group_perf}, we show the race-wise performance in five trials.
Given a certain race $s$, the sequence of $\{\textrm{ADR}_s(k)\}_{k\in[N]}$ for one trial forms a time series. Across all five trials, the mean value and $\pm $ one standard deviation could be calculated from the five time series.
We denote the mean value of the time series across five trials by a solid curve and $\pm $ one standard deviation by error shades, with the corresponding race distinguished by colour.

In Figures~\ref{fig:indiv_perf_ts} and~\ref{fig:indiv_perf_heatmap}, we show the user-wise performance in five trials.
Similarly, given a certain user $i$, the sequence of $\{\textrm{ADR}_i(k)\}_{k\in[N]}$ for one trial is a time series. 
From the five trials, and all users in $[N]$, $1000\times 5$ time series.
%, are visualised in both Figures~\ref{fig:indiv_perf_ts} and~\ref{fig:indiv_perf_heatmap}.
In Figure~\ref{fig:indiv_perf_ts}, the $1000\times 5$ time series are visualised directly, with their races distinguished by colours.
In Figure~\ref{fig:indiv_perf_heatmap}, the race information of the users are erased, as we intend to present the distribution of the $1000\times 5$ time series by grey shades.
Note that darker shades denote higher density of ADR$_i(k)$ at the certain time step.

Recalling the goal of equal impact in \eqref{eq:goal-ADR}, we would like to see these time series converge (weakly to the same distribution).
From Figure~\ref{fig:group_perf}-\ref{fig:indiv_perf_heatmap}, we do observe that all time series, aggregated by race or not, are dwindling to a similar level.
%Figure~\ref{fig:indiv_perf_ts} displays the time series $\{\textrm{ADR}_i(k)\}_{k\in[N]}$ for all users from 5 trials ($5\times 1000$ curves), with their races distinguished by colour.
%In Figure~\ref{fig:indiv_perf_heatmap}, the race information is ignored, and the grey shading shows the density of ADR$_i(k)$ at different time steps.

%\textcolor{red}{we should say that the goal is the ADR somewhere.}

\section{Conclusions}

We have presented a novel, closed-loop view of the impact of AI systems. 
On the example in consumer-credit approvals, we showcase, that equal impact is possible while preserving equal treatment conditional on a non-protected attribute of income. 
An important question for further work is how to impose constraints on the equality of impact \cite{celis2019classification}.
Another important question asks whether the coupling arguments of Hairer et al. \cite{Hairer2011} could make it possible to show certain contrapositive statements, suggesting when such guarantees are impossible to provide. 

%\section*{Acknowledgments}
%Quan's and Bob's work has been supported by the Science Foundation Ireland under Grant 16/IA/4610. Jakub acknowledges 
%support of 
%the OP RDE funded project CZ.02.1.01/0.0/0.0/16\_019/0000765 ``Research Center for Informatics''.
%This work has received funding from the European Union’s Horizon Europe research and innovation programme under grant agreement No. 101070568. This work was also supported by Innovate UK under the Horizon Europe Guarantee; UKRI Reference Number: 10040569 (Human-Compatible Artificial Intelligence with Guarantees (AutoFair)).

%\FloatBarrier
\bibliographystyle{apalike}
%\bibliographystyle{plain} 
% \bibliography{main} 
%\end{document}

%\clearpage
%\textcolor{red}{up to eight (8) pages plus one more page containing only references.}

\clearpage
\appendix
\section{Markov Systems}

\begin{figure}[ht]
\centering
\includegraphics[width=0.48\textwidth]{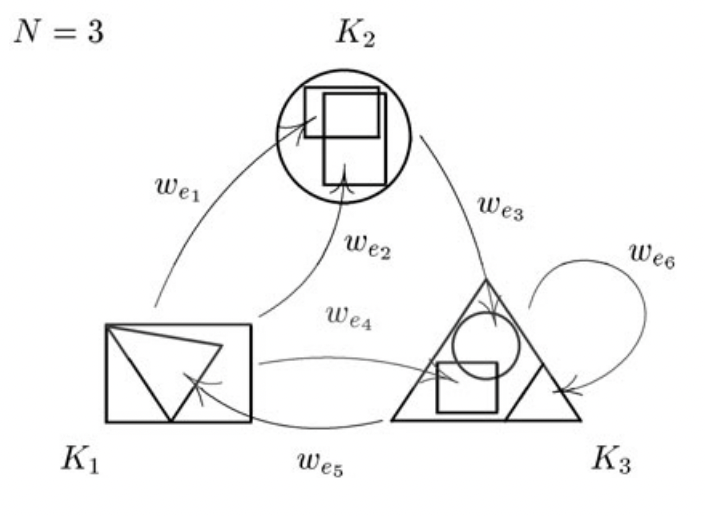}
\caption{A Markov system \cite{Ivan2004}}
\label{fig:ms}
\end{figure}

A Markov system (see Figure~\ref{fig:ms}) is a family $\left(X_{i(e)}, w_e, p_e\right)_{e\in E}$ where $E$ consisting of edges of a finite directed (multi) graph $\left(V, E, i, t\right)$ with $V=\{1,2,\dots,N\}$ are vertices and $N=1$ is also possible, $i:E\to V$ indicates the initial vertex of each edge and $t:E\to V$ indicates the terminal vertex of each edge, $X_1,\dots, X_N$ is a partition of the metric space $(X,d)$ into non-empty Borel subsets, $\left(w_{e}\right)_{e\in E}$ is a family of Borel-measurable maps on the metric space such that 
\begin{align*}
    w\left(X_{i(e)}\right)\subseteq X_{t(e)}\;\text{for all } e\in E,
\end{align*}
and $\left(p_{e}\right)_{e\in E}$ is a family of Borel measurable maps on $X$ with the property $p_e(x)\ge 0$ for all $e\in E$ and $\sum\limits_{e\in E} p_e(x)=1$\; for all $x\in X$. A Markov system is called irreducible or aperiodic if its directed graph is irreducible or aperiodic. A Markov system is called contractive with contraction factor $a$ if its probability functions satisfy the following average contractivity condition, $\text{for all } x,y\in X_i,\; i=1,2,\dots, N$,
\begin{align*}
    \sum\limits_{e\in E}p_e(x)d(w_e(x), w_e(y))\le a d(x,y).
\end{align*}
The Markov system defined above determines a Markov operator $\mathrm{P}$ on the space of bounded Borel measurable functions on $X$, which is denoted by $\mathcal{L}^0(X)$,
\begin{align*}
\mathrm{P}f(x)=\sum\limits_{e\in E} p_e f\circ w_e\; \text{ for all } f\in \mathcal{L}^0(X),
\end{align*}
and the adjoint of $\mathrm{P}$ is denoted by $\mathrm{P}^{\star}$ acts on the space of Borel probability measures $\mathcal{M}_{p}(X)$ as
\begin{align*}
  \mathrm{P}^{\star}\nu(f)=\int \mathrm{P}(f) \mathrm{d}\nu\; \text{for all } \nu\in   \mathcal{M}_{p}(X).
\end{align*}
A Borel probability measure $\mu$ is said to be an invariant probability measure for the Markov system if it is a stationary distribution of the associated Markov process i.e.
\begin{align*}
    \mathrm{P}^{\star}\mu=\mu.
\end{align*}
A Borel probability measure $\mu$ is called attractive for the contractive Markov system iff 
\begin{align*}
    \lim_{n\to \infty}(\mathrm{P}^{\star})^n\nu\to \mu\; \text{ for all } \nu\in \mathcal{M}_{p}(X).
\end{align*}

\section{Incremental Stability}

Incremental stability is a well-established concept to describe the asymptotic property of differences between any two solutions. One can utilise the concept of incremental input-to-state stability, which is defined as follows:

\begin{definition}\label{dfn:ch5-class-K}
A function $\gamma : \mathbb R^{+}\to \mathbb R^{+}$ is is said to be of class $\mathcal K$ if it is continuous, increasing and $\gamma(0)=0$. It is of class $\mathcal K_{\infty}$ if, in addition, it is proper, i.e., unbounded.
\end{definition}

\begin{definition}\label{dfn:ch5-class-KL}
A continuous function $\beta: \mathbb R^{+}\times \mathbb R^{+}\to \mathbb R^{+}$ is said to be of class $\mathcal K \mathcal L$, if for all fixed $t$ the function $\beta(\cdot, t)$ is of class $\mathcal K$ and for all fixed $s$, the function $\beta(s, \cdot)$ is is non-increasing
and tends to zero as $t\to \infty$.
\end{definition}

\begin{definition}[Incremental ISS, \cite{Angeli2002}]
Let $\mathcal U$ denote the set of all input functions $u: \mathbb Z_{\ge k_0}\to \mathbb R^d$
Suppose $F: \mathbb R^d \times \mathbb R^n\to \mathbb R^n$ is continuous, then the discrete-time non-linear dynamical system
\begin{align}
x(k+1) = F(x(k),u(k)),
\label{system-inciss}
\end{align}
is called (globally) \emph{incrementally input-to-state-stable} (incrementally ISS), if there exist $\beta\in \mathcal{KL}$ and $\gamma\in\mathcal K$ such that
for any pair of inputs $u_1, u_1\in\mathcal{U}$ and any pair of initial condition  $\xi_1, \xi_2 \in \mathbb R^n$:
\begin{align*}
\|x(k, \xi_1, u_1)-x(k, \xi_2, u_2)\|
\le \beta(\|\xi_1 - \xi_2\|, k)+\\ \gamma(\|u_1-u_2\|_{\infty}),\quad \forall k\in \mathbb N. 
\end{align*}
\end{definition}

\end{document}